\documentclass{article}
\usepackage[hyphens]{url}
\usepackage{spconf,amsmath,graphicx}
\usepackage[nolist]{acronym}
\usepackage{amssymb}
\usepackage{tikz}
\usetikzlibrary{positioning, calc}
\usepackage{makecell}
\usepackage{multirow}
\usepackage{booktabs}
\usepackage{nicefrac}
\usepackage[linesnumbered,ruled,vlined]{algorithm2e}
\usepackage{physics}
\usepackage{enumitem}

\begin{acronym}
\acro{ASR}{automatic speech recognition}
\acro{CTC}{connectionist temporal classification}
\acro{AED}{attention-based encoder decoder}
\end{acronym}

\title{Neural transducer training: reduced memory consumption with sample-wise computation}
\name{Stefan Braun, Erik McDermott and Roger Hsiao}
\address{Apple \\ \texttt{\{stefan\_braun, erik\_mcdermott, rhsiao\}@apple.com}}

\begin{document}
\ninept
\maketitle
\begin{abstract}
The neural transducer is an end-to-end model for \ac{ASR}. While the model is well-suited for streaming ASR, the training process remains challenging. During training, the memory requirements may quickly exceed the capacity of state-of-the-art GPUs, limiting batch size and sequence lengths. 

In this work, we analyze the time and space complexity of a typical transducer training setup. We propose a memory-efficient training method that computes the transducer loss and gradients sample by sample. We present optimizations to increase the efficiency and parallelism of the sample-wise method. In a set of thorough benchmarks, we show that our sample-wise method significantly reduces memory usage, and performs at competitive speed when compared to the default batched computation. As a highlight, we manage to compute the transducer loss and gradients for a batch size of 1024, and audio length of 40 seconds, using only 6 GB of memory.

\end{abstract}
\section{Introduction}
\label{sec:intro}
End-to-end models are compelling approaches for \ac{ASR} due to their simplicity in model architecture and training process. A number of end-to-end models have been proposed in the literature, such as \ac{CTC} \cite{Graves2006ConnectionistTC}, neural transducer \cite{Graves2012SequenceTW} and \ac{AED} models \cite{Chorowski2015AttentionBasedMF, Chan2016ListenAA}. 
Due to its convenient properties for streaming \ac{ASR}, the neural transducer has seen increased research interest over the past years \cite{Battenberg2017ExploringNT, Rao2017ExploringAD, Zhang2020TransformerTA}, resulting in successful deployment in on-device production settings \cite{He2019StreamingES}. However, training a transducer model is typically more complex and resource intensive than training CTC or \ac{AED} models. Recent efforts have focused on improving performance and memory consumption of transducer training \cite{warp-transducer, Li2019ImprovingRT, Zhang2021BenchmarkingLC, Kuchaiev2019NeMoAT, Bagby2018EfficientIO, kuang2022pruned}.

During transducer training, five components are involved, shown in Fig. \ref{fig:transducer_training}: acoustic encoder, label encoder, joint network, output layer and loss function. The output layer computes a 4D tensor of shape $[B, T, U, V]$, with batch size $B$, acoustic sequence length $T$, label sequence length $U$, and vocabulary size $V$. Even for moderate settings, the memory size of the 4D output tensor can quickly grow into the \textbf{10s to 100s of GBs} - exceeding even the largest GPU memory configurations \footnote{V100: up to 32GB, A100/H100: up to 80GB}.

To reduce memory consumption, previous work has proposed various optimizations. Computational optimizations use function merging to avoid intermediate results tensors \cite{Graves2012SequenceTW, warp-transducer, Li2019ImprovingRT}, and reuse memory by overwriting activation tensors in-place with gradient tensors \cite{Li2019ImprovingRT}. Further memory savings are achieved by methods that reduce the memory size of the 4D output tensor: reduced precision \cite{Zhang2021BenchmarkingLC}, pruning \cite{kuang2022pruned}, padding removal \cite{Li2019ImprovingRT} and batch-splitting \cite{Kuchaiev2019NeMoAT}. 

The batch-splitting method in the NVIDIA NeMo toolkit \cite{Kuchaiev2019NeMoAT} computes the acoustic encoder for the full batch size, and then splits the remainder of the computation into sub-batches. After the gradients for a sub-batch are computed, the occupied memory can be freed before moving on to the next sub-batch. The full batch gradients are obtained by gradient accumulation.

This work extends previous investigations with a full analysis for the time and space complexity of transducer training. To save memory, we propose a \textit{sample-wise} method that is similar, but not identical to batch-splitting. Our implementation maximizes the amount of batched computation, and minimizes memory consumption in the remainder of the computation. We propose a novel method to dynamically increase the parallelism during sample-wise computation. We provide a set of thorough performance and memory benchmarks, using parameter settings that are typical for real-world training.

\section{Transducer training}

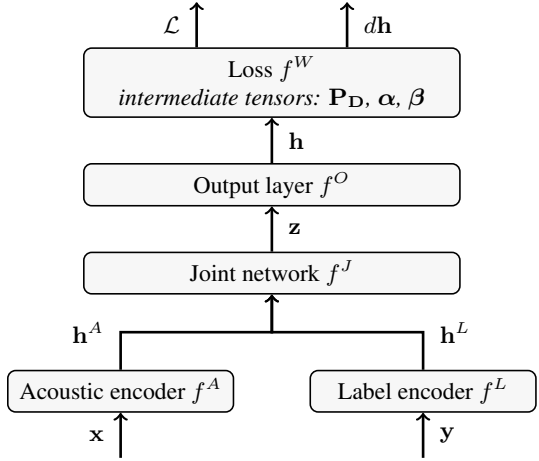
\begin{figure}
    \tikzset{every picture/.style={line width=1pt}}
    \centering
    \begin{tikzpicture}[node distance=1cm and 1cm,
        rectnode/.style={rectangle, draw=black!100, fill=black!3, semithick, minimum width=30mm, rounded corners=3},
        rectnodewide/.style={rectangle, draw=black!100, fill=black!3, semithick, minimum width=50mm,rounded corners=3},
        ]
        \node[rectnode](ae){Acoustic encoder $f^A$};
        \node[rectnode](le)[right=of ae] {Label encoder $f^L$};
        \node[rectnodewide](jn)[above=of $(le.north)!0.5!(ae.north)$] {Joint network $f^J$};
        \node[rectnodewide](ol)[above=0.6cm of jn]{Output layer $f^O$};
        \node[rectnodewide, align=center](lo)[above=0.6cm of ol]{Loss $f^W$\\\textit{intermediate tensors: $\mathbf{P_D}$, $\boldsymbol\alpha$, $\boldsymbol\beta$}};
        
        \draw[<-] (ae.south) to node[left=0.1cm] {$\mathbf{x}$} ++(0.,-0.6cm);
        \draw[<-] (le.south) to node[right=0.1cm] {$\mathbf{y}$} ++(0.,-0.6cm);
        \draw[->] (ae.north) |- ++(0.,0.5cm) node[left=0.1cm] {$\mathbf{h}^A$} -| (jn.south);
        \draw[->] (le.north) |- ++(0.,0.5cm) node[right=0.1cm] {$\mathbf{h}^L$} -| (jn.south);
        \draw[->] (jn.north) to node[right=0.1cm]{$\mathbf{z}$} (ol.south);
        \draw[->] (ol.north) to node[right=0.1cm]{$\mathbf{h}$} (lo.south);
        \draw[->] ($(lo.north)+(1.0,0cm)$) to node[right=0.1cm]{$d \mathbf{h}$} ++(0,.6cm);
        \draw[->] ($(lo.north)+(-1.0,0cm)$) to node[left=0.1cm]{$\mathcal{L}$} ++(0,.6cm);
\end{tikzpicture}
\caption{The transducer training process.}
\label{fig:transducer_training}
\end{figure}

This section describes the transducer training process as depicted in Fig. \ref{fig:transducer_training}. We assume that the transducer loss is based on the open-source implementation \texttt{warp-transducer} \cite{warp-transducer}. The same or similar loss implementations are used in \ac{ASR} toolkits such as TorchAudio \cite{yang2021torchaudio}, ESPnet \cite{watanabe2018espnet} and NeMo \cite{Kuchaiev2019NeMoAT}. The following description is meant to provide sufficient detail for the complexity analysis in Section \ref{sec:complexity_intro}; for a formal description of the overall transducer model, the reader is referred to existing literature \cite{Graves2012SequenceTW, warp-transducer}.

Transducer training involves five components: (1) the acoustic encoder $f^A$, (2) the label encoder $f^L$, (3) the joint network $f^J$, (4) the output layer $f^O$ and (5) the loss function $f^W$. The training inputs consist of the acoustic features tensor $\mathbf{x}$ with shape ${[B, T, F]}$, and the labels tensor $\mathbf{y}$ with shape ${[B, U+1]}$. The dimensions represent batch size $B$, acoustic sequence length $T$, acoustic features dimension $F$, and labels sequence length $U$. In our implementation, the labels tensor is prepended with the blank token $\emptyset$, producing the dimensionality $U+1$.

The acoustic encoder $f^A$ with parameters $\theta^A$ converts the acoustic features into the acoustic encodings $\mathbf{h}^{A}$, a tensor with shape $[B, T, H_A]$ and acoustic encoding size $H_A$ (Eq. (\ref{eq:fa})).
\begin{equation}
    \mathbf{h}^A = f^{A}(\mathbf{x})
    \label{eq:fa}
\end{equation}

The label encoder $f^L$ with parameters $\theta^L$ converts the training labels into the label encodings $\mathbf{h}^L$ , a tensor with shape $[B, U+1, H_L]$ and label encoding size $H_L$ (Eq.(\ref{eq:fl})).
\begin{equation}
    \mathbf{h}^L = f^L(\mathbf{y})
    \label{eq:fl}
\end{equation}
The joint network $f^J$ then combines the acoustic encodings and label encodings into the joint encodings $\mathbf{z}$, a tensor with shape $[B, T, U+1, H]$ and joint encoding size $H$ (Eq. (\ref{eq:z})). The joint encodings $\mathbf{z}$ at indices $[b,t,u]$ are defined as $\mathbf{z}_{b,t,u} \in \mathbb{R}^H$ (Eq. (\ref{eq:zbtu})). The parameters of the joint network consist of $\theta^J=\{\mathbf{W}^A \in \mathbb{R}^{H\times H_A}, \mathbf{W}^L \in \mathbb{R}^{H\times H_L}, \mathbf{b}^Z \in \mathbb{R}^H \}$.
\begin{align}
    \mathbf{z} &= f^J(\mathbf{h}^A, \mathbf{h}^L) \label{eq:z}\\
    \mathbf{z}_{b,t,u} &= \tanh (\textstyle \mathbf{W}^A \mathbf{h}^A_{b,t} + \mathbf{W}^L\mathbf{h}^L_{b,u}+\mathbf{b}^Z) \label{eq:zbtu}
\end{align}
The output layer $f^O$ projects the joint encodings to output scores $\mathbf{h}$, a tensor with shape $[B, T, U+1, V]$ and vocabulary size $V$ (Eq. (\ref{eq:h})). The output scores $\mathbf{h}$ at indices $[b,t,u]$ are defined as $\mathbf{h}_{b,t,u} \in \mathbb{R}^V$ (Eq. (\ref{eq:hbtu})). The parameters of the output layer consist of $\theta^O = \{\mathbf{W^O} \in \mathbb{R}^{V\times H}, \mathbf{b}^O \in \mathbb{R}^V\}$.
\begin{align}
    \mathbf{h} &= f^O(\mathbf{z}) \label{eq:h}\\
    \mathbf{h}_{b,t,u} &= \mathbf{W}^O\mathbf{z}_{b,t,u} + \mathbf{b}^O \label{eq:hbtu}
\end{align}

Finally, the overall loss function $f^W$ is split into 4 sub-functions $f^W_{1-4}$, computing the softmax denominator $\mathbf{P}_D$ with shape $[B,T,U+1]$; the forward and backward variables $(\boldsymbol{\alpha}, \boldsymbol{\beta})$, both with shape $[B,T,U+1]$; the scalar loss value $\mathcal{L}$; and the output score gradients $d\mathbf{h}=\nicefrac{\partial \mathcal{L}}{\partial \mathbf{h}}$ with shape $[B, T, U+1, V]$:
\begin{align}
\mathbf{P}_D& = f^W_1(\mathbf{h}) \label{eq:fw1}\\
\boldsymbol{\alpha, \beta}& = f^W_{2}(\mathbf{h}, \mathbf{P}_D)\\
\mathcal{L}& = f^W_{3}(\boldsymbol{\alpha}) \label{eq:fw3}\\
d\mathbf{h}& = f^W_{4}(\mathbf{h}, \mathbf{P}_D, \boldsymbol{\alpha}, \boldsymbol{\beta})
\end{align}

The \texttt{warp-transducer} loss reduces memory consumption by introducing step $f^W_1$ in Eq. (\ref{eq:fw1}). When normalizing the output scores to probabilities, naively computing the softmax would result in a 4D probability tensor $\mathbf{P}$ with shape $[B, T, U+1, V]$. Instead, only the softmax denominator $\mathbf{P}_D$ is computed, which is $V$ times smaller. The computation of actual probabilities is deferred to an efficient in-place op in steps $f^W_{2,4}$. The loss $\mathcal{L}$ in Eq. (\ref{eq:fw3}) corresponds to $\mathcal{L}= - \sum_{b=1}^B \ln P(\mathbf{y}_b|\mathbf{x}_b)$; for more details, see \cite{Graves2012SequenceTW, warp-transducer}.

\section{Time and space complexity}
\label{sec:complexity_intro}
The time and space complexities for the transducer functions are given in Table \ref{tb:complexity}. The analysis covers the computation of the joint network $f^J$, output layer $f^O$ and loss function $f^W$ as defined in Section 2. The computations for the acoustic encoder $f^A$ and label encoder $f^L$ are omitted, as their complexity depends on the user's choice of model size and architecture.

\begin{table}[]
    \centering
    \caption{Time \& space complexities for transducer training components. For space complexity, the proposed \texttt{sample-wise } method reduces the dimension $B$ to 1.}
    \vspace{1mm}
    \label{tb:complexity}
\begin{tabular}{ lrrr }
\toprule 
& \multicolumn{2}{c}{Complexity} &\\
\cmidrule(lr){2-3}
Function & Space & Time\\ 
\midrule
$f^J$ & $BTUH$ &  $BTUH (1+\frac{H_A}{U}+\frac{H_L}{T})$\\
$f^O$ & $BTU(H+V)$ & $BTUHV$ \\
$f^W$ & $BTUV$ & $BTUV$\\
\midrule
$f^W_1$ &$BTUV$ & $BTUV$\\
$f^W_{2}$ & $BTUV$ & $BU(T+U)$\\
$f^W_{3}$ & $BTU$ & $B$ \\
$f^W_{4}$ & $BTUV$ & $BTUV$ \\
\bottomrule
\end{tabular}
\end{table}
\vspace{-2mm}

\subsection{Time complexity}
\label{sec:complexity_time}
The time complexities for the joint network $f^J$, output layer $f^O$ and softmax denominator function $f^W_1$ are based on standard neural network components such as linear layers, pointwise operations and the softmax op. For the forward-backward algorithm in $f_2^W$ \cite{Graves2012SequenceTW}, the \texttt{warp-transducer} loss uses an efficient dynamic programming algorithm that is best described in the source code \cite{warp-transducer}. The sub-functions for loss $f^W_3$ and output score gradients $f^W_4$ are computed by combining previous results for $\mathbf{h}, \mathbf{P_D}, \boldsymbol\alpha$ and $\boldsymbol\beta$.

By counting the number of factors in the time complexity in Table \ref{tb:complexity}, functions are ranked from highest to lowest complexity:
\begin{itemize}
    \item 5 factors: output layer $f^O$;
    \item 4 factors: joint network $f^J$, softmax denominator $f^W_1$, output score gradients $f^W_4$;
    \item 3 factors: forward-backward algorithm $f^W_2$;
    \item 1 factor: loss value $f^W_3$.
\end{itemize}
\vspace{-2.5mm}

\subsection{Space complexity}
\label{sec:complexity_space}
The space complexity covers both input space and auxiliary space, therefore including input tensors and output tensors. Between $f^J$, $f^O$ and $f^W$, all functions operate on 4D tensors, and therefore have similar space complexity. The main differentiator is the relative size of the hidden dimension $H$ vs. the vocabulary size $V$.
\vspace{-2.mm}

\subsection{Complexity reduction and practical relevance}
\label{sec:complexity_potential}

Given the use of standard neural network ops, and efficient dynamic programming algorithms, there is limited potential to reduce time and space complexity beyond reducing dimensions $B,T,U,H,H_A,H_L,V$. While the $f^J$, $f^O$ or $f^W$ may be formulated in alternative ways to reduce the number of ops, this may negatively affect model quality and is beyond the scope of this work. Finally, even the reduction of dimensions has its limitations: it is only the batch size $B$ that can be reduced without limiting input lengths or model size. 

From a practical point of view, the time complexity of transducers may not be a critical issue. The computations with highest time complexity, e.g. in the output layer, are highly parallel and ideal targets for specialized tensor processing accelerators (TPU \cite{jouppi2021ten}, TensorCores \cite{tensorcores}). However, space complexity remains a critical issue, as the size of the 4D tensors ($\mathbf{z}, \mathbf{h}, d\mathbf{h}$) may exceed memory capacity.

\section{Sample-wise transducer training}
\label{sec:gradient accumulation}

This section describes a sample-wise method to make transducer training more memory efficient. Memory is saved by reducing the effective batch size to $B=1$ for select transducer components.

The method is detailed in Algorithm \ref{alg:gradient_accumulation}. The notation $[b]$ refers to indexing into position $b$ of the batch dimension. A tensor with the star superscript $(^*)$ corresponds to a single training sample, i.e. the tensor is not batched. For increased readability, the notation $dx$ refers to the partial derivative of the loss w.r.t. the corresponding variable $x$, i.e. $dx=\nicefrac{\partial\mathcal{L}}{\partial x}$. The variables $d\boldsymbol{\theta}^{J,O,A,L}$ correspond to the gradients w.r.t. to the parameters of the joint network, output layer, acoustic encoder and label encoder, respectively.

First, the acoustic and label encoders run in default batched mode, producing batched encodings (lines 1, 2). Then, the computation switches to a sample-wise mode for the joint network, output layer and loss function (lines 4-11). Both activations and gradients are computed with a batch size of 1, saving memory. When the computation for an individual sample is finished, the gradients are accumulated (lines 8-11). Due to the accumulation, the result is identical to batched computation. By deleting the intermediate tensors $(\mathbf{z}^*, \mathbf{h}^*, d\mathbf{h}^*)$, memory is freed (line 12). The remaining downstream gradients for the acoustic and label encoder parameters are computed in default batched mode (lines 13, 14).

\begin{algorithm}
\label{alg:gradient_accumulation}
\DontPrintSemicolon
\tcp{Batched computation}
\BlankLine
$\mathbf{h}^A \gets f^A(\mathbf{x})$ \tcp*{Acoustic encodings}
$\mathbf{h}^L \gets f^L(\mathbf{y})$ \tcp*{Label encodings}
\BlankLine
\tcp{Sample-wise computation}
\BlankLine
$d\boldsymbol{\theta}^J, d\boldsymbol{\theta}^O, d\mathbf{h}^A, d\mathbf{h}^L, \mathcal{L} \gets 0$ \tcp*{Init. variables}

\BlankLine
\For{b $\gets 1$ \KwTo $B$}{
$\mathbf{z}^* \gets f^J(\mathbf{h}^A[b], \mathbf{h}^L[b])$ \tcp*{Process sample} 

$\mathbf{h}^*  \gets f^O(\mathbf{z}^*)$ 

$\mathcal{L}[b], d\mathbf{h}^* \gets f^W(\mathbf{h}^*)$ 

$d\boldsymbol{\theta}^O \gets d\boldsymbol{\theta}^O + d\mathbf{h}^*\frac{\partial\mathbf{h}^*}{\partial\boldsymbol{\theta}^O}$

$d\boldsymbol{\theta}^J \gets d\boldsymbol{\theta}^J + d\mathbf{h}^*\frac{\partial\mathbf{h}^*}{\partial\boldsymbol{\theta}^J}$

$d\mathbf{h}^A[b] \gets d\mathbf{h}^*\frac{\partial\mathbf{h}^*}{\partial\mathbf{h}^A}$

$d\mathbf{h}^L[b] \gets d\mathbf{h}^*\frac{\partial\mathbf{h}^*}{\partial\mathbf{h}^L}$

Delete $\mathbf{z}^*,\mathbf{h}^*,d\mathbf{h}^*$ \tcp*{Release memory}
}
\tcp{Batched computation}
$d\boldsymbol{\theta}^A \gets d\mathbf{h}^A \frac{\partial\mathbf{h}^A}{\partial\mathbf{\boldsymbol{\theta}}^A}$

$d\boldsymbol{\theta}^L \gets d\mathbf{h}^L \frac{\partial\mathbf{h}^L}{\partial\mathbf{\boldsymbol{\theta}}^L}$

\BlankLine

\caption{Sample-wise transducer training}
\end{algorithm}
\vspace{-5mm}

\subsection{Optimizations}

\subsubsection{Padding removal}
When generating training batches, samples are typically selected with a \textit{bucketing} approach. With bucketing, samples of similar acoustic sequence length are batched to reduce wasted computation on padded segments. However, the transducer combines acoustic and label encodings. The variance in label sequence lengths can lead to a significant amount of padding e.g. in the joint network and output layer. The sample-wise computation allows us to remove all padding, fully avoiding wasted computation.

\subsubsection{Dynamic parallelism}
The proposed method partially reduces the batch size to $B=1$, which reduces efficiency on parallel accelerators. However, some of the parallelism can be recovered by running multiple iterations of the sample-wise computation in parallel (lines 4-12). In our TensorFlow implementation, we use the \texttt{tf.while\_loop} with a dynamic setting for \texttt{parallel\_iterations}. Based on the size of the intermediate tensor  $\textbf{h}^*$, the parallelism is dynamically controlled to stay within user-defined memory limits. For short batches with small $\mathbf{(T,U)}$, higher parallelism can be afforded without exceeding memory limits. For longer batches with large $\mathbf{(T,U)}$, \texttt{parallel\_iterations} will gradually drop to 1.

\subsection{Related work}

The NVIDIA NeMo toolkit proposes a similar batch-splitting method to save memory with neural transducers \cite{Kuchaiev2019NeMoAT}. The computation of the label encoder, joint network, output layer and loss is split into sub-batches with reduced batch size $>=1$. The sub-batch size is fixed, and only the acoustic encoder runs in batched mode.
In contrast, our proposed method runs both acoustic and label encoders in batched mode, increasing parallelism from batching. The rest of the model runs in a sample-wise mode, corresponding to batch size $B=1$. The parallelism in the sample-wise computation is dynamically adapted by adjusting the number of parallel iterations based on the size of $(T,U)$. In \cite{kuang2022pruned}, an intermediate estimation step is used to select (T,U) positions for pruning, saving memory. 
Our approach uses sample-wise computation without pruning.


\section{Benchmarks}
\subsection{Setup}
The benchmarks measure the memory usage and performance of transducer training. The evaluation covers the computation of activations and gradients for three components: joint network, output layer and transducer loss. The acoustic and label encoders are omitted, as they depend on user choice. The acoustic and label encodings $(\mathbf{h}^A, \mathbf{h}^L)$ are simulated by randomly initializing tensors with a fixed seed before starting the benchmarks. During the benchmarks, the pre-initialized tensors are provided as input to the joint network. The benchmarks are implemented in TensorFlow 2.8 and executed on a A100 GPU with 40 GB of memory. Three warmup steps are followed by 100 benchmark steps. The performance is reported as the median over benchmark steps. The peak memory usage is obtained with \texttt{tf.config.experimental.get\_memory\_info()}.

Four algorithms are benchmarked: (1) \texttt{batched} - the default baseline implementation; (2) \texttt{sample-wise} - the sample-wise method proposed in Section \ref{sec:gradient accumulation}; (3) \texttt{sample-wise +PR} - adding padding removal; and (4) \texttt{sample-wise +PR +DP} - adding both padding removal and dynamic parallelism. When using dynamic parallelism, the number of parallel iterations is computed from a range $\texttt{PI}=1..16$ based on the size of $T,U,V$ (Eq. (\ref{eq:pi})). 
\begin{align}
    \texttt{PI}=2^{\max(0 , \min(4, \lfloor \log(\nicefrac{10^9}{4TUV}) / \log(2) \rfloor))}
    \label{eq:pi}
\end{align}

The parameter settings are chosen to reflect real-world training setups. The hidden dimension $H=1024$ and vocabulary size $V=4096$ are fixed throughout all the benchmarks. The batch size is varied between $B=1..1024$, and input lengths are varied between $T=50..500$ and $U=10..100$. With a typical frame rate of 100 fps for feature extraction, and 8x downsampling in the acoustic encoder, this would correspond to audio lengths $T'$ from 4 to 40 seconds.

To emulate real-world variable-length input, zero-padding is simulated. For acoustic encodings, zero padding linearly increases from 0\% (first sample in batch) to 9.3\% (last sample in batch). For label encodings, zero-padding increases similarly from 0\% to 45.8\%\footnote{zero-padding percentages estimated from real-world trainings using data bucketing based on acoustic features lengths}. This guarantees that there is always a sample without zero-padding in the batch, provoking maximum memory usage.

\subsection{Memory}

\begin{figure}
    \centering
    \includegraphics[width=\columnwidth]{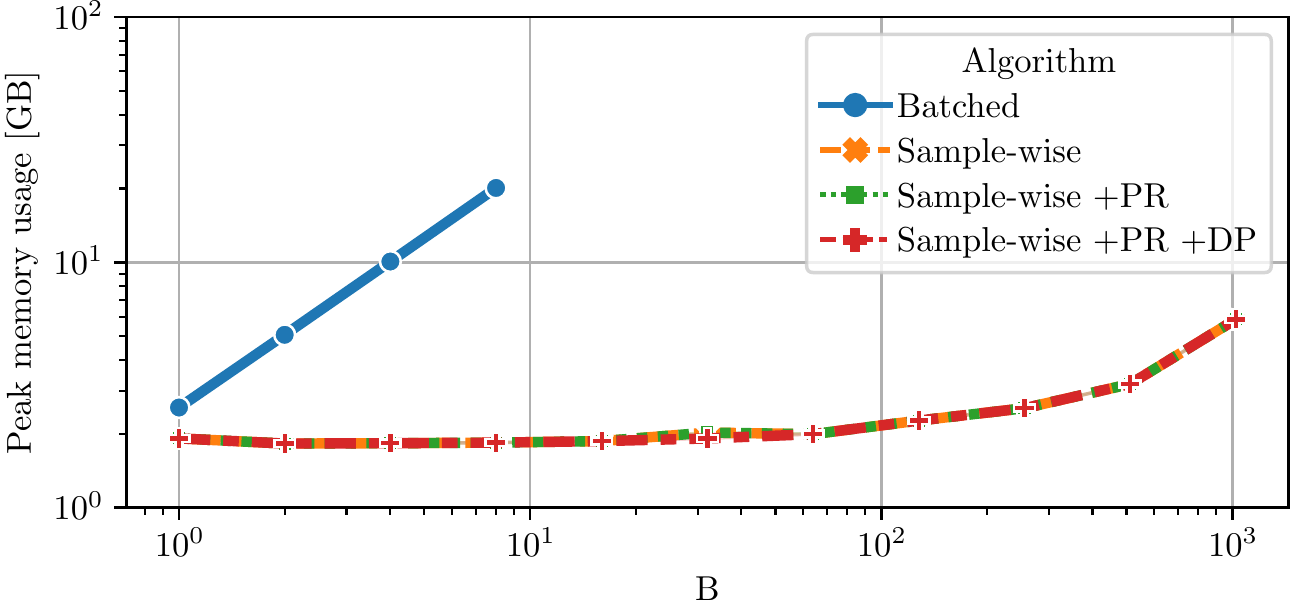}
    \vspace*{-7.0mm}
    \caption{Peak memory usage over batch size.}
    \label{fig:memory_over_b}
\end{figure}

The first benchmark varies the batch size between $B=1..1024$, and fixes input lengths to the maximum values $T \times U = 500 \times 100$. Results are given in Fig. \ref{fig:memory_over_b}. The \texttt{sample-wise} methods consume less than 6 GB of memory for $B=1024$. Notably, the \texttt{sample-wise} methods make the memory usage almost independent from batch size between $B=1..256$. In contrast, the default \texttt{batched} method runs out-of-memory for batch sizes $B\geq16$.

\begin{figure}
    \centering
    \includegraphics[width=\columnwidth]{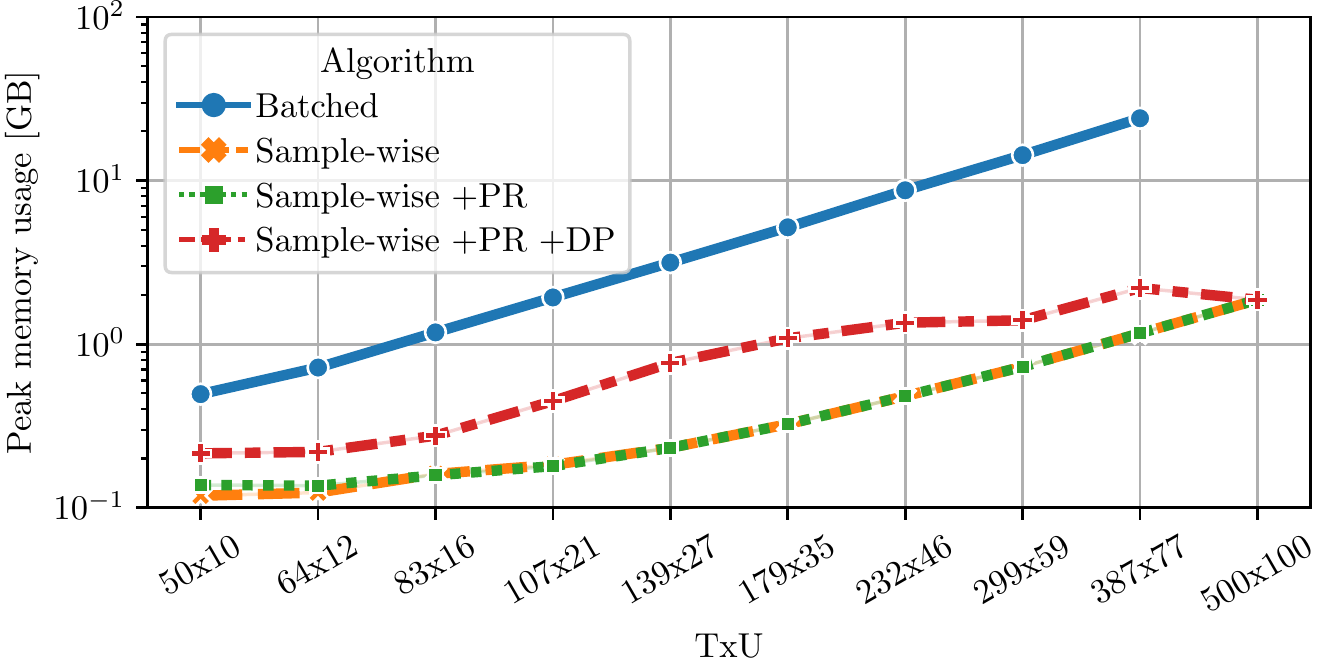}
    \vspace*{-7mm}
    \caption{Peak memory usage over input lengths.}
    \label{fig:memory_over_txu}
\end{figure}

The second benchmark fixes the batch size to $B=16$ and varies input lengths, with results given in Fig. \ref{fig:memory_over_txu}. The \texttt{sample-wise} and \texttt{sample-wise +PR} methods have the lowest memory usage over all $T\times U$, consuming at most 1.86 GB of GPU memory. The \texttt{sample-wise +PR +DP} method has slightly increased memory usage, as multiple samples may be processed at the same time. The \texttt{batched} method has the highest memory usage, running out of memory for $T \times U=500\times100$. 

\subsection{Performance}
\begin{figure}
    \centering
    \includegraphics[width=\columnwidth]{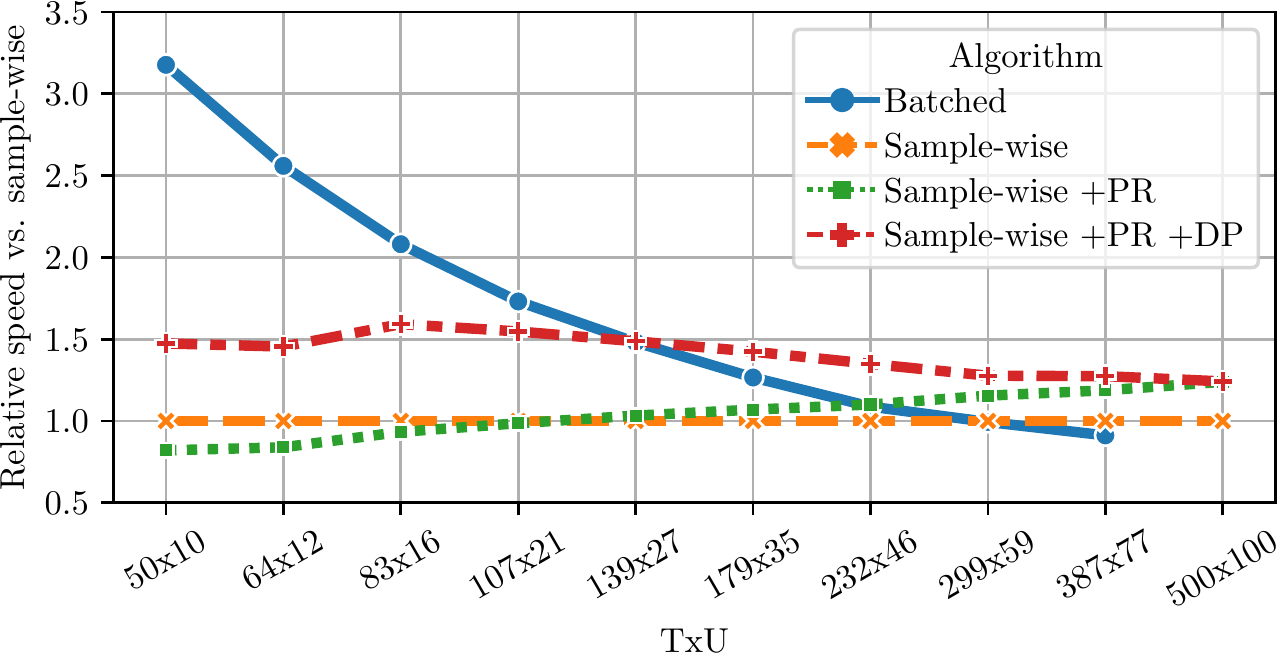}
    \vspace*{-7mm}
    \caption{Speed relative to the \texttt{sample-wise} method.}
    \label{fig:performance_over_txu}
\end{figure}

The performance benchmark measures the step time for $B=16$ and $T\times U=50\times10..500\times100$. The speed is normalized w.r.t. the \texttt{sample-wise} method and results are shown in Fig. \ref{fig:performance_over_txu}. Compared to \texttt{sample-wise}, the \texttt{batched} method reaches relative speed between 0.91x to 3.18x. The speed advantage is most prominent for short input lengths, and both \texttt{batched} and \texttt{sample-wise} converge to similar speeds beyond $T \times U=232  \times 46$. This is explained by the size of the output layer computation: for longer input lengths, the output layer can saturate the GPU even for single samples. Therefore, the benefit of batching becomes negligible.

The method \texttt{sample-wise +PR} removes zero-padding. Compared to \texttt{sample-wise}, relative speed is 0.82x to 1.24x. The speed loss occurs at short input lengths due to the overhead of the slicing op. When adding dynamic parallelism, \texttt{sample-wise +PR +DP} performs 1.24x to 1.59x faster than \texttt{sample-wise}.

Overall, the fastest method depends on the $T \times U$ setting. Notably, for input lengths with $T \times U \geq 139 \times 27$, the memory-efficient method \texttt{sample-wise +PR +DP} performs the best.

\section{Conclusion}
This work presented a sample-wise method to reduce the memory consumption for the computation of transducer loss and gradients. The computation in joint network, output layer and transducer loss is carried out with a batch size of 1. As a result, the memory consumption is almost independent of batch size: we were able to compute the transducer gradients on large batch sizes of up to 1024, and audio lengths of 40 seconds, within less than 6 GB of memory.

The sample-wise processing allows the removal of all padding from individual samples, saving computation. While the sample-wise method comes with an inherent loss of parallelism, some parallelism is recovered by using the parallel for loop in TensorFlow. Here we proposed to adapt the number of parallel iterations dynamically, depending on the input size. This allows one to balance memory consumption and performance, as befits a specific compute/resource environment. 

Future work may further improve memory consumption and/or performance by combining sample-wise computation with other techniques such as mixed precision techniques and pruning.

\section{Acknowledgements}
We would like to thank Dogan Can, Xinwei Li, Ruoming Pang, Awni Hannun and Russ Webb for their support and useful discussions.

\vfill\pagebreak
\bibliographystyle{IEEEbib}
\bibliography{strings,refs}

\end{document}